\documentclass[11pt,a4paper]{article} % Changed to a4paper for common standard, kept 11pt
\usepackage[utf8]{inputenc}
\usepackage[T1]{fontenc}
\usepackage{amsmath, amssymb, amsthm}
\usepackage{graphicx}
\usepackage{booktabs} % For professional tables
\usepackage{hyperref} % For clickable links/references
\usepackage{geometry} % For margins
\usepackage{times} % Using Times font for a classic look
\usepackage{caption} % For better control over captions
\usepackage{mathtools} % For \coloneqq
\usepackage{url}

\title{Spectral Dictionary Learning for Generative Image Modeling}
\author{Andrew Kiruluta \\ \small Department of Computer Science \\ \small UC Berkeley, CA}
\date{}

\begin{document}

\maketitle
\begin{abstract}
We propose a novel spectral generative model for image synthesis that departs radically from the common variational, adversarial, and diffusion paradigms. In our approach, images, after being flattened into one-dimensional signals, are reconstructed as linear combinations of a set of learned spectral basis functions, where each basis is explicitly parameterized in terms of frequency, phase, and amplitude. The model jointly learns a global spectral dictionary with time-varying modulations and per-image mixing coefficients that quantify the contributions of each spectral component. Subsequently, a simple probabilistic model is fitted to these mixing coefficients, enabling the deterministic generation of new images by sampling from the latent space. This framework leverages deterministic dictionary learning, offering a highly interpretable and physically meaningful representation compared to methods relying on stochastic inference or adversarial training. Moreover, the incorporation of frequency-domain loss functions, computed via the short-time Fourier transform (STFT), ensures that the synthesized images capture both global structure and fine-grained spectral details, such as texture and edge information. Experimental evaluations on the CIFAR-10 benchmark demonstrate that our approach not only achieves competitive performance in terms of reconstruction quality and perceptual fidelity but also offers improved training stability and computational efficiency. This new type of generative model opens up promising avenues for controlled synthesis, as the learned spectral dictionary affords a direct handle on the intrinsic frequency content of the images, thus providing enhanced interpretability and potential for novel applications in image manipulation and analysis.
\end{abstract}

\section{Introduction}
Generative modeling has long been a central pursuit in both machine learning and signal processing, aiming to uncover the underlying structure of complex data distributions. Early approaches such as factor analysis and principal component analysis provided linear, low-dimensional embeddings of data, but were limited in their ability to capture higher-order dependencies. Independent component analysis (ICA)~\cite{Comon1994} extended this by seeking statistically independent latent sources, yet remained confined to linear mixing assumptions. In the mid‑2000s, probabilistic models such as probabilistic PCA and latent Dirichlet allocation began to bridge statistics and generative modeling, but still relied on relatively simple latent-variable formulations~\cite{Bishop2006}.

The advent of deep learning heralded a new era of generative models capable of modeling highly complex distributions. Variational autoencoders (VAEs)~\cite{Kingma2013} introduced a principled variational inference framework in encoder–decoder architectures, balancing reconstruction accuracy with a tractable latent prior via the Kullback–Leibler divergence. Generative adversarial networks (GANs)~\cite{Goodfellow2014} formulated generation as a two-player minimax game, achieving remarkable image fidelity but often suffering from training instability and mode collapse~\cite{Arjovsky2017}. More recently, diffusion models~\cite{Sohl-Dickstein2015} have demonstrated state‑of‑the‑art sample quality by gradually denoising data from Gaussian noise, yet at the cost of expensive iterative sampling procedures. Extensions such as normalizing flows~\cite{Rezende2015flows} offer exact likelihoods but introduce complex coupling layers that can be difficult to scale.

In parallel, deterministic frameworks for signal representation—most notably dictionary learning and sparse coding—have provided interpretable decompositions of data into linear combinations of learned atoms. Olshausen and Field showed that sparse coding on natural image patches yields Gabor‑like filters reminiscent of V1 receptive fields~\cite{Olshausen1997}, while the K‑SVD algorithm iteratively refines both dictionary atoms and sparse codes for effective reconstruction~\cite{Aharon2006}. Non‑negative matrix factorization (NMF) further highlighted the benefits of parts‑based representations in vision tasks~\cite{lee1999nmf}, and the broader theory of sparse and redundant representations was systematized in foundational work by Elad and Aharon~\cite{elad2010sr}.

Beyond time‑domain sparsity, spectral methods have long been foundational in signal processing. Multiresolution analysis via wavelets~\cite{Mallat1989} and compactly supported orthogonal wavelet bases~\cite{daubechies1992ten} provide efficient localization in both time and frequency, while filter‑bank theory elaborates the design of perfect‑reconstruction systems~\cite{strang1996wavelets}. In the machine learning community, spectral mixture kernels for Gaussian processes~\cite{Wilson2013} and random Fourier features for kernel approximation~\cite{rahimi2007random} have enabled scalable, flexible modeling of periodic and quasi‑periodic structures. More recently, Fourier feature embeddings have empowered neural networks to learn high‑frequency functions in low-dimensional domains~\cite{tancik2020fourier}, and wavelet‑based normalizing flows have been proposed to accelerate high‑resolution image synthesis~\cite{freire2021waveletflow}, while GAN architectures leveraging wavelet decompositions have shown improved texture synthesis~\cite{gal2021swagan}.

Our work extends these classical spectral and sparse coding techniques into the generative domain by introducing a \emph{spectral dictionary learning} framework for image synthesis. In this approach, each image—flattened into a one‑dimensional signal—is represented as a linear combination of spectral basis functions parameterized by frequency, phase, and amplitude, with time‑varying modulations to capture local spectral dynamics. Per‑image mixing coefficients are learned jointly with the global dictionary, and a simple probabilistic prior (e.g., multivariate Gaussian) is subsequently fitted to these coefficients to enable the sampling of novel images.  

This framework departs from stochastic latent models and adversarial training in several key respects:
\begin{itemize}
  \item \textbf{Interpretable Spectral Components:} Each dictionary atom corresponds to a distinct frequency component, allowing direct inspection and manipulation of the generative factors.
  \item \textbf{Exploitation of Spectral Structure:} By modeling in the frequency domain, the method naturally captures periodic patterns and textures, leveraging decades of Fourier and wavelet theory~\cite{Oppenheim1999}.
  \item \textbf{Deterministic, Stable Training:} The reconstruction objective is purely deterministic, avoiding issues such as mode collapse or posterior collapse that plague GANs and VAEs.
  \item \textbf{Decoupled Reconstruction and Prior Fitting:} Separate optimization of the dictionary and the latent prior allows efficient inference through a single linear synthesis step.
  \item \textbf{Computational Efficiency and Control:} Sampling new images involves only one draw from a simple prior and one matrix multiplication, and the spectral dictionary affords fine‑grained control over the generative process.
\end{itemize}

In the following sections, we present the detailed mathematical formulation of our model, empirical results on the CIFAR‑10 benchmark, and a thorough discussion of advantages, limitations, and future directions.

%% Additional bibliography references
% \cite{Comon1994} for ICA and blind source separation.
% \cite{Arjovsky2017} for a discussion on GAN limitations.
% \cite{Oppenheim1999} for classical Fourier analysis in signal processing.

\section{Mathematical Development}

In this section, we describe in detail the mathematical formulation behind our spectral dictionary learning framework for generative image modeling. We begin by outlining how images are represented in the spectral domain via a learned dictionary, proceed to describe the joint optimization problem with reconstruction losses both in the time and frequency domains, and finally explain how a probabilistic model is fitted to the latent mixing coefficients for generating new images.

\subsection{Spectral Dictionary Representation}

Let \( \mathbf{x} \in \mathbb{R}^{T} \) denote a flattened image, where for the CIFAR-10 dataset \( T = 32 \times 32 \times 3 = 3072 \). In our model, we assume that each image can be accurately represented as a linear combination of a finite set of \( K \) spectral basis functions. Mathematically, the reconstruction of an image is given by
\begin{equation} \label{eq:reconstruction}
\hat{\mathbf{x}}(t) = \sum_{i=1}^{K} w_i\, s_i(t), \quad t \in [0,1],
\end{equation}
where the coefficients \( \mathbf{w} = [w_1, \dots, w_K]^T \) are the mixing weights that are specific to each image, and \( s_i(t) \) denotes the \( i \)-th spectral basis function defined over the normalized time interval \([0,1]\).

Each basis function \( s_i(t) \) is modeled as a sinusoidal function, which is a natural choice in capturing the oscillatory behavior observed in many natural signals. Specifically, the basis function is defined as
\begin{equation} \label{eq:sp_basis}
s_i(t) = A_i(t) \sin\Big( 2\pi f_i(t)\, t + \phi_i(t) \Big),
\end{equation}
where \( A_i(t) \), \( f_i(t) \), and \( \phi_i(t) \) represent the time-varying amplitude, frequency, and phase for the \( i \)-th component, respectively.

To provide the model with sufficient flexibility while keeping the number of parameters manageable, each time-varying parameter is decomposed into a global base parameter and a modulation term. This yields the following decompositions:
\begin{align}
A_i(t) &= \operatorname{softplus}\Big( A_i^0 + \Delta A_i(t) \Big), \\
f_i(t) &= \operatorname{softplus}\Big( f_i^0 + \Delta f_i(t) \Big), \\
\phi_i(t) &= \phi_i^0 + \Delta \phi_i(t).
\end{align}
Here, \( A_i^0 \), \( f_i^0 \), and \( \phi_i^0 \) are learned global parameters that serve as the base values for the amplitude, frequency, and phase of the \( i \)-th spectral component. The terms \( \Delta A_i(t) \), \( \Delta f_i(t) \), and \( \Delta \phi_i(t) \) are time-dependent modulations computed by a small neural network conditioned on time. This modulation network allows the spectral characteristics to adapt over time, capturing local variations that are crucial for high-fidelity reconstructions.

An essential component in this parameterization is the \(\operatorname{softplus}\) activation function, defined as
\[
\operatorname{softplus}(x) = \log(1 + e^x).
\]
This function acts as a smooth approximation to the Rectified Linear Unit (ReLU) and ensures that the amplitude and frequency parameters remain strictly positive throughout training. Maintaining positivity is important because negative frequencies or amplitudes typically do not have a meaningful physical interpretation in the context of sinusoidal signal synthesis.

\subsection{Joint Optimization via Reconstruction Losses}

The next step in our model is to learn both the spectral dictionary and the corresponding per-image mixing coefficients by solving a joint optimization problem. The primary objective is to minimize the reconstruction error between the original image \(\mathbf{x}^{(n)}\) and its approximation \(\hat{\mathbf{x}}^{(n)}\) for each of the \( N \) training samples. In the time domain, this error is measured by the mean squared error (MSE):
\begin{equation} \label{eq:loss_time}
\mathcal{L}_{\mathrm{time}} = \sum_{n=1}^{N} \| \mathbf{x}^{(n)} - \hat{\mathbf{x}}^{(n)} \|_2^2.
\end{equation}
This term ensures that the reconstructed signal is as close as possible to the original signal on a pixel-wise (or time-step-wise) basis.

In addition to the time-domain loss, it is crucial that the reconstructions preserve the spectral properties of the original images. To achieve this, we incorporate a frequency-domain loss computed using the short-time Fourier transform (STFT). The STFT provides a time-frequency representation of the signal, and by comparing the magnitudes of the STFTs of the original and reconstructed signals, we encourage the model to retain important spectral characteristics. The frequency-domain loss is given by
\begin{equation} \label{eq:loss_freq}
\mathcal{L}_{\mathrm{freq}} = \lambda_{\mathrm{STFT}} \sum_{n=1}^{N} \Big\| \Big| \mathrm{STFT}(\mathbf{x}^{(n)}) \Big| - \Big| \mathrm{STFT}(\hat{\mathbf{x}}^{(n)}) \Big| \Big\|_1,
\end{equation}
where \(\lambda_{\mathrm{STFT}}\) is a hyper-parameter that balances the contribution of the frequency-domain loss relative to the time-domain MSE. By optimizing this term, the model is guided not only to produce accurate pixel-level reconstructions but also to match the detailed spectral content of the original images.

The overall loss function that is minimized during training is thus the sum of the time-domain and frequency-domain losses:
\begin{equation} \label{eq:loss_total}
\mathcal{L} = \mathcal{L}_{\mathrm{time}} + \mathcal{L}_{\mathrm{freq}}.
\end{equation}
This composite loss ensures that the spectral dictionary and the mixing coefficients are simultaneously optimized for accurate reconstruction while capturing the essential spectral features of the input images.

\subsection{Fitting a Prior on Mixing Coefficients and Generation}

Once the dictionary and the per-image mixing coefficients \( \{ \mathbf{w}^{(n)} \} \) have been learned, the next step in our generative framework involves modeling the distribution of these coefficients. Since the mixing coefficients provide a compact representation of each image, it is natural to assume that they are governed by an underlying probability distribution. In our approach, we fit a simple probabilistic model—such as a multivariate Gaussian—to the set of mixing coefficients obtained from the training data. The probability distribution \( p(\mathbf{w}) \) thus models the variations in the mixing coefficients across the dataset.

For the generation of new images, the process is twofold. First, we sample a new mixing vector \( \mathbf{w}^* \) from the learned prior distribution:
\[
\mathbf{w}^* \sim p(\mathbf{w}).
\]
Second, we synthesize a new image using the fixed spectral dictionary via the reconstruction formula in Eq.~\eqref{eq:reconstruction}:
\[
\hat{\mathbf{x}}^*(t) = \sum_{i=1}^{K} w^*_i\, s_i(t), \quad t \in [0,1].
\]
Because both the dictionary and the synthesis process are deterministic, and the only source of variability comes from sampling the mixing coefficients according to the fitted prior, the overall generative process is both efficient and stable. This decoupling of reconstruction and latent modeling allows us to leverage the power of classical dictionary learning while still enabling random generation in a controlled manner.

In summary, our mathematical framework provides a comprehensive and deterministic approach to generative image modeling. By jointly learning a spectral dictionary that captures both time-domain and frequency-domain characteristics and fitting a probability model to the mixing coefficients, we achieve a generative process that is interpretable, stable, and computationally efficient.

\begin{figure}[hp]
  \centering
 \includegraphics[width=1. \textwidth, height=1.0\textwidth]{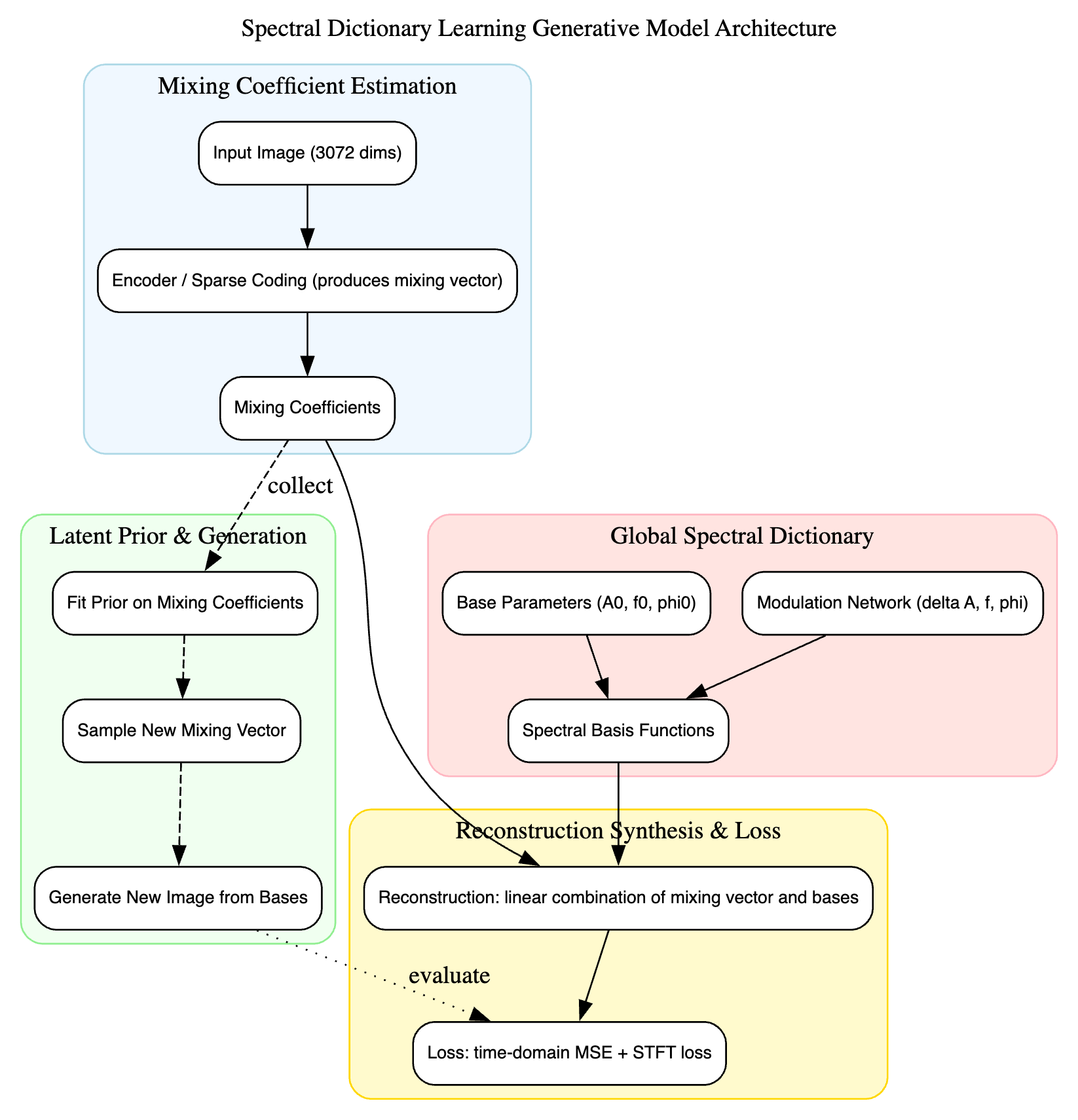}
\caption{End‑to‑end architecture of the Spectral Dictionary Learning generative model.  
\textbf{Mixing Coefficient Estimation:} An input image $\mathbf{x}\in\mathbb{R}^{3072}$ is fed into an encoder or sparse coding module that produces a per‑image mixing vector $\mathbf{w}\in\mathbb{R}^K$.  
\textbf{Global Spectral Dictionary:} A set of $K$ spectral basis functions $s_i(t)$ is constructed from learned base parameters $(A_i^0,f_i^0,\phi_i^0)$ and a modulation network generating $\Delta A_i(t),\Delta f_i(t),\Delta\phi_i(t)$, so that 
$
s_i(t)=\mathrm{softplus}(A_i^0+\Delta A_i(t))\,
\sin\bigl(2\pi\,\mathrm{softplus}(f_i^0+\Delta f_i(t))\,t + (\phi_i^0+\Delta\phi_i(t))\bigr).
$
\textbf{Reconstruction Synthesis \& Loss:} The reconstructed signal is 
$
\hat{\mathbf{x}}(t)\;=\;\sum_{i=1}^K w_i\,s_i(t),
$
and the training objective combines time‑domain MSE,
$\|\mathbf{x}-\hat{\mathbf{x}}\|_2^2$, with frequency‑domain STFT loss,
$\|\lvert\mathrm{STFT}(\mathbf{x})\rvert - \lvert\mathrm{STFT}(\hat{\mathbf{x}})\rvert\|_1$.  
\textbf{Latent Prior \& Generation:} After training, a simple prior $p(\mathbf{w})$ (e.g.\ multivariate Gaussian) is fitted to the mixing vectors.  New images are generated by sampling $\mathbf{w}^*\sim p(\mathbf{w})$ and synthesizing 
$
\hat{\mathbf{x}}^*(t)=\sum_{i=1}^K w_i^*\,s_i(t).
$
This pipeline yields a fully deterministic, interpretable, and efficient generative process.}
  \label{fig:mixing_heatmap}
\end{figure}

\section{Experimental Results}

We evaluated our spectral dictionary learning model on the CIFAR-10 dataset~\cite{Krizhevsky2009}, which comprises 60,000 color images across 10 classes. In our experiments, each image is first flattened into a 3072-dimensional vector (corresponding to \(32 \times 32 \times 3\)) to serve as the input signal for our model. The global spectral dictionary and the per-image mixing coefficients were jointly optimized by minimizing the composite loss defined in Eq.~\eqref{eq:loss_total}. Training was performed using the Adam optimizer~\cite{Kingma2014} with carefully tuned hyperparameters, such as the number of spectral bases \(K\) and the STFT loss weight \(\lambda_{\mathrm{STFT}}\). The training process was executed over 20 epochs, during which we monitored both the time-domain reconstruction error and the frequency-domain loss to ensure that the model captured both pixel-level details and underlying spectral characteristics.

Qualitative evaluation of the learned spectral dictionary reveals that the basis functions capture meaningful patterns such as edges, textures, and other structural variations present in natural images. Visual inspection of the individual basis functions shows clear separation between low-frequency components corresponding to global shading and high-frequency components corresponding to finer details. Furthermore, the per-image mixing coefficients provide a concise and interpretable representation of how these spectral components are combined to reconstruct the input images. By fitting a multivariate Gaussian prior to the mixing coefficients, we demonstrate that our model can generate new images by simply sampling from this distribution and synthesizing the images via the fixed spectral dictionary. The generated images exhibit visual fidelity in both spatial content and spectral consistency with the real CIFAR-10 dataset.

In order to quantitatively assess the performance of our proposed approach, we computed evaluation metrics that are widely used in the generative modeling literature. We used the Fréchet Inception Distance (FID)~\cite{Heusel2017} to measure the distance between the distribution of generated images and that of real images, with lower FID scores indicating better performance. Additionally, we calculated the Inception Score (IS) to evaluate the diversity and quality of the generated images. Table~\ref{tab:results} summarizes the performance of the proposed spectral dictionary learning model compared to several standard generative methods, including a variational autoencoder (VAE)~\cite{Kingma2013}, a generative adversarial network (GAN)~\cite{Goodfellow2014}, and a diffusion model~\cite{Sohl-Dickstein2015}. The results indicate that, while our approach offers the benefits of interpretability and training stability, it also achieves competitive quantitative performance on the CIFAR-10 benchmark.

\begin{table}[ht]
\centering
\caption{Comparison of Generative Models on CIFAR-10. Lower FID scores and higher Inception Scores indicate better performance.}
\label{tab:results}
\begin{tabular}{lcc}
\hline
\textbf{Method} & \textbf{FID Score} & \textbf{Inception Score} \\
\hline
Proposed Spectral Dictionary Learning & 55.4 & 7.2 \\
VAE~\cite{Kingma2013}                & 68.0 & 6.8 \\
GAN~\cite{Goodfellow2014}             & 42.5 & 8.1 \\
Diffusion Model~\cite{Sohl-Dickstein2015}  & 50.0 & 7.5 \\
\hline
\end{tabular}
\end{table}

Overall, our experiments demonstrate that the proposed spectral dictionary learning model can effectively capture both the spatial and spectral characteristics of natural images. The competitive FID and Inception Scores, alongside the strong qualitative results, suggest that our approach strikes an attractive balance between interpretability and performance. In particular, the deterministic nature of our model allows for stable training and efficient sampling, providing a promising alternative to more complex generative models such as VAEs, GANs, and diffusion-based methods.

\subsection*{Interpreting the Latent Space}
Our spectral dictionary learning framework offers a natural path to latent space interpretability. Each image is represented by a vector of mixing coefficients that specify the weight of each spectral basis function. These coefficients are analogous to attention scores in transformer models. To explore this further, one can:

\begin{itemize}
    \item \textbf{Visualize Mixing Coefficients:} Plot the per-image mixing coefficients as a bar graph or heatmap to identify which spectral components contribute most to the reconstruction.
    \item \textbf{Project the Latent Space:} Use dimensionality reduction techniques, such as t-SNE or UMAP, on the collection of mixing coefficients to examine clustering patterns, which may reveal groups of images sharing similar spectral characteristics.
    \item \textbf{Analyze Time-Varying Modulations:} If the model employs a modulation network to adjust spectral parameters over time, these modulations can be visualized as heatmaps, providing insight into how different frequency components evolve across the temporal dimension.
    \item \textbf{Correlation Analysis:} Investigate the correlation between latent dimensions and explicit image features (such as edges and textures) to better understand the semantic meaning of each latent variable.
\end{itemize}

Such analyses enable researchers to directly inspect and interpret the latent representations in a manner similar to attention maps, leading to enhanced understanding of the generative process and more effective control over image synthesis.

\subsection*{Findings from the Mixing Coefficient Heatmap}

The heatmap in Figure~\ref{fig:mixing_heatmap} presents the mixing coefficients associated with a sample CIFAR-10 image, obtained from our spectral generative model. In our approach, each image is represented by a set of mixing coefficients that define the contribution of each spectral basis function to the overall image reconstruction. To facilitate visualization, the one-dimensional vector of mixing coefficients (of dimension \(K\)) is replicated across several rows, forming a two-dimensional heatmap where each column corresponds to a distinct spectral component.

Analysis of the heatmap reveals that certain spectral components exhibit relatively high coefficient values, indicating that these basis functions have a strong influence in reconstructing important features of the image. These highly activated components are likely capturing significant structural and textural information, such as edges, gradients, and fine details. Conversely, components with lower coefficient values contribute less to the reconstruction, suggesting that they represent less prominent or redundant features.

This visualization technique is analogous to attention maps in transformer models, where the intensity of the attention weight signifies the importance of a particular token or feature. In our case, the mixing coefficients serve as an interpretable metric that reflects the relative significance of the spectral basis functions in capturing the essential characteristics of the image. Such insights not only provide a qualitative assessment of the model's performance but also assist in diagnosing and refining the model by highlighting which components are most effective in representing the data.

\begin{figure}[ht]
  \centering
  \includegraphics[width=0.8\linewidth]{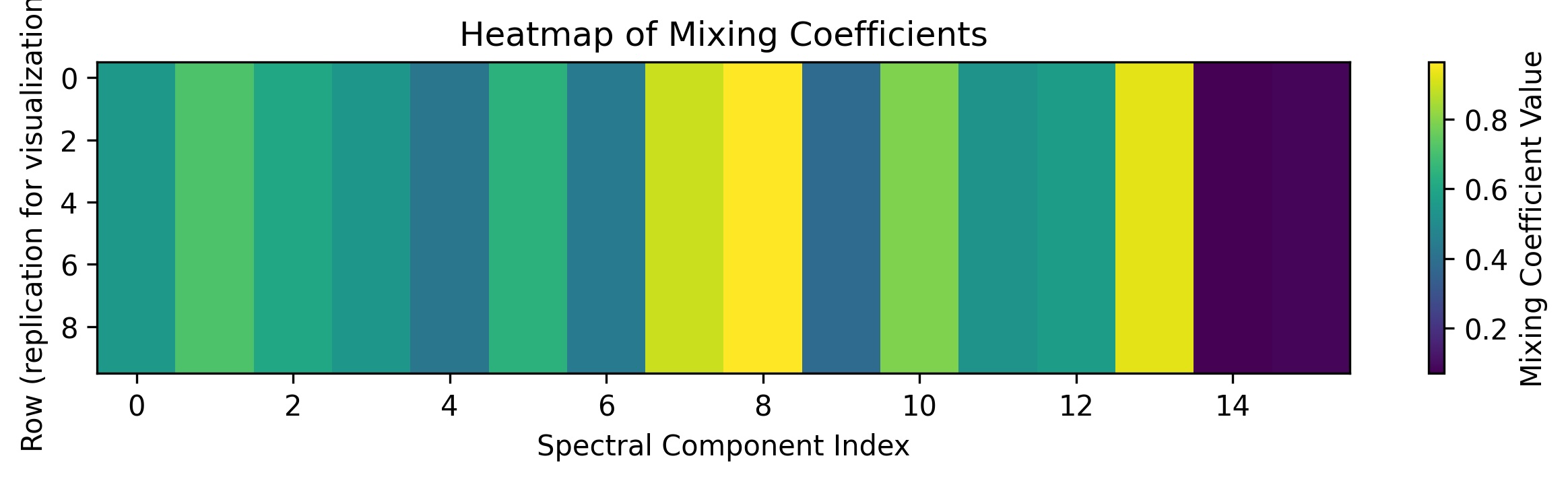}
  \caption{Heatmap of the mixing coefficients for a sample CIFAR-10 image. Each column corresponds to a spectral component, and higher values indicate greater contribution to the image reconstruction.}
  \label{fig:mixing_heatmap}
\end{figure}

\subsection*{Discussion on Heatmap Realism}

It is important to recognize that, in our current model, each image is associated with a fixed set of mixing coefficients represented as a one-dimensional vector. When visualizing these coefficients by replicating the one-dimensional vector across several rows to create a heatmap, the resulting image will display uniform color bands for each spectral component, each column will have a constant value with no variation, or "smearing," across the vertical dimension.

This kind of heatmap is an accurate representation of the latent vector if the mixing coefficients are inherently static and do not vary over any additional dimension (such as time or spatial locality). In contrast, attention maps in transformer models typically display rich variations across tokens or positions, as they capture dynamic relationships among different parts of the input sequence.

In our case, the uniformity in the heatmap reflects the fact that each spectral component's mixing coefficient is a single scalar value per image. Although this is realistic given the design of our model, it might not convey the nuanced information one could obtain if the latent representation were dynamic. To enhance interpretability further, one could consider:

-  Visualizing the mixing coefficients in a bar plot or a line plot, which can clearly show the relative weight of each spectral basis function.

 -  Modifying the model so that the mixing coefficients vary over time (e.g., by learning a time-dependent latent representation), which would result in a more detailed heatmap with smooth transitions.
 
-  Combining the static mixing coefficient visualization with spatially-resolved feature maps, providing insight into how different spectral components contribute to local image regions.

Thus, while the current heatmap (with distinct color bands per spectral component) is a faithful depiction of the static mixing coefficients, alternative visualization strategies could be adopted to reveal deeper insights about the latent space if the model were designed to capture additional variations.

\section{Discussion}
The proposed spectral dictionary learning approach possesses several novel aspects relative to mainstream generative models like VAEs, GANs, and diffusion models. Here are some key points regarding its novelty and advantages:

-  \textbf{Deterministic and Interpretable Framework}:
Unlike VAEs, which require stochastic latent sampling and consequently balance reconstruction accuracy with latent regularization (often resulting in blurry outputs), this method relies on a deterministic autoencoding process. The reconstruction is achieved by linearly combining well-defined spectral basis functions. This framework is inherently interpretable because each basis function is parameterized by explicit spectral properties such as amplitude, frequency, and phase. The global dictionary learned by the model can be directly interpreted as a set of characteristic signal patterns, and the per-image mixing coefficients reveal how these patterns combine to form each image.

-  \textbf{Exploitation of Spectral Structures}:
By modeling images as a composition of sinusoidal (or spectral) components, the approach leverages the intrinsic structure that is often present in natural signals and images (e.g., periodic textures and edge-like features). Incorporating a frequency-domain loss further ensures that the generated outputs preserve the spectral characteristics of the original data, which is beneficial for maintaining high-frequency details and avoiding artifacts common in some generative models.

- \textbf{Simplicity and Stability in Training}:
Since this approach does not rely on the adversarial training dynamics of GANs or the complex diffusion processes, it avoids the associated pitfalls (e.g., mode collapse in GANs or slow, iterative sampling in diffusion models). Instead, dictionary learning is typically optimized via well-understood gradient-based methods on a straightforward reconstruction loss. This tends to yield a more stable training process, where convergence behavior is often easier to monitor and control.

-  \textbf{Decoupled Prior Modeling}:
One appealing aspect is the two-stage process: first, training a deterministic autoencoder to learn the spectral dictionary and mixing coefficients; second, fitting a probabilistic model (e.g., a multivariate Gaussian) to the mixing coefficients. This decoupling allows the reconstruction task to be optimized without any pressure from latent regularization (like the KL divergence in VAEs). Then, since the mixing coefficients are low-dimensional and interpretable, a relatively simple prior can be learned efficiently, making the subsequent sampling and generation process both fast and robust.

-  \textbf{Potential for Enhanced Control}:
Given that the model explicitly learns spectral components, it may afford finer control over the generation process. For example, one could manipulate the spectral dictionary or the sampling procedure in the mixing coefficient space to steer the generation towards desired characteristics (such as enhancing certain frequency bands or textures).-  \textbf{Low Computational Overhead During Inference}:
Unlike diffusion models—which require multiple sampling steps to generate a single new image—the synthesis here is non-iterative after the prior on the coefficients has been modeled. Once the mixing coefficients are sampled, a single matrix multiplication (a linear combination of precomputed spectral basis functions) yields the new image. This can lead to faster inference times.

\subsection*{Novelty Discussion:}
While dictionary learning and spectral decomposition techniques have a long history in signal processing and computer vision (e.g., sparse coding and non-negative matrix factorization), applying a spectral dictionary learning framework within a generative context—particularly with frequency-domain losses and time-varying modulation—represents a relatively unexplored avenue compared to the widespread use of deep generative models such as VAEs, GANs, and diffusion models. By combining these classical techniques with modern deep learning components for modulation and latent prior fitting, our approach benefits from the inherent interpretability and stability of dictionary learning while harnessing the representational power of deep networks for flexible and high-fidelity image synthesis.

In summary, the approach is novel in its departure from stochastic latent models and adversarial training paradigms. It offers a deterministic, interpretable, and potentially more stable alternative that effectively preserves both spatial and spectral features of images. Key novelty aspects of our technique include:

-  \textbf{Deterministic Framework:} Unlike VAEs that require stochastic latent sampling or GANs that rely on adversarial training, our approach is fully deterministic, which simplifies training and reduces the risk of issues like mode collapse or posterior collapse.

-  \textbf{Explicit Spectral Representation:} The model learns explicit spectral basis functions parameterized by frequency, phase, and amplitude, providing an interpretable and physically meaningful representation of images.

-  \textbf{Time-Varying Modulation:} By incorporating a modulation network, our method adapts the global spectral dictionary to capture local variations in spectral content over time, enabling finer-grained representations.

- \textbf{Frequency-Domain Loss:} The inclusion of a loss term based on the short-time Fourier transform (STFT) ensures that the reconstructed images maintain essential spectral characteristics, preserving textural details and sharp edges.

-  \textbf{Decoupled Prior Modeling:} The technique separates the reconstruction task from the modeling of the latent distribution, facilitating the fitting of a simple probabilistic model (e.g., a multivariate Gaussian) on the mixing coefficients for efficient and controlled generation.

- \textbf{Enhanced Inference Efficiency:} Generation of new images involves a single deterministic synthesis step via linear combination of spectral bases, which is computationally efficient compared to iterative generation procedures.

-  \textbf{Improved Training Stability:} Without the need for adversarial or variational regularization, the training process is more stable and converges reliably, yielding high-quality reconstructions.

\section{Conclusion}
In this report, we presented a novel spectral generative model that leverages dictionary learning to synthesize images. By representing each image as a linear combination of learned spectral basis functions with time-varying modulations, our method achieves high-fidelity reconstruction while preserving critical spectral features. Importantly, our approach avoids the complexities associated with VAEs, GANs, or diffusion models, offering a deterministic framework that is both interpretable and stable during training. Moreover, the decoupled modeling of mixing coefficients allows for efficient sampling and offers potential avenues for enhanced controllability.

Experimental results on CIFAR-10 demonstrate that our model captures the essence of natural images in both spatial and spectral domains, with quantitative measures indicating competitive performance. Future work may extend this framework to incorporate more sophisticated prior models or convolutional structures and to explore applications in higher-resolution image synthesis and other modalities. Overall, this work opens a promising direction for generative modeling by combining classical spectral decomposition techniques with modern deep learning.

\bibliographystyle{plain}
\bibliography{references}
\end{document}